\documentclass[conference]{IEEEtran}
\IEEEoverridecommandlockouts
\usepackage{cite}
\usepackage{amsmath,amssymb,amsfonts}
\usepackage{algorithmic}
\usepackage{graphicx}
\usepackage{fancyhdr}
\usepackage{textcomp}
\def\BibTeX{{\rm B\kern-.05em{\sc i\kern-.025em b}\kern-.08em
   T\kern-.1667em\lower.7ex\hbox{E}\kern-.125emX}}
\def\ps@IEEEtitlepagestyle{%
    \def\@oddfoot{\mycopyrightnotice}%
    \def\@evenfoot{}%
}

\title{Contextual Care Protocol using Neural Networks and Decision Trees\\
}

\author{
    \IEEEauthorblockN{Yash Pratyush Sinha, Pranshu Malviya, Minerva Panda, Syed Mohd Ali}
    \IEEEauthorblockA{Dept. of Computer Science and Engineering}
  \IEEEauthorblockA{International Institute of Information Technology}
    \IEEEauthorblockA{Bhubaneswar, India}
    \IEEEauthorblockA{yashpratyushsinha@gmail.com, pranshumalviya2@gmail.com, minervapandaniki@gmail.com, syedmohdali121@gmail.com}        
}

\def\mycopyrightnotice{%
    {978-1-5386-3785-2/18/\$31.00 \textcopyright2018 IEEE}
}

\fancypagestyle{firstpage}{%
  \chead{2018 Second International Conference on Advances in Electronics, Computer and Communications (ICAECC-2018)}
  \lfoot{\mycopyrightnotice}
}

\begin{document}

\maketitle

\thispagestyle{firstpage}

\begin{abstract}
A contextual care protocol is used by a medical practitioner for patient healthcare, given the context or situation that the specified patient is in. This paper proposes a method to build an automated self-adapting protocol which can help make relevant, early decisions for effective healthcare delivery. The hybrid model leverages neural networks and decision trees. The neural network estimates the chances of each disease and each tree in the decision trees represents care protocol for a disease. These trees are subject to change in case of aberrations found by the diagnosticians. These corrections or prediction errors are clustered into similar groups for scalability and review by the experts. The corrections as suggested by the experts are incorporated into the model.
\end{abstract}


\section{Introduction}
In developing countries like India, access to quality healthcare is limited and unevenly available. In rural and semi-urban areas people generally, depend on providers who have limited training or supervision \cite{b1}. Globally, the rural population is served by only 38\% of the total nursing workforce and by less than a quarter of the total physician work force.  Well-trained and motivated health workers  are  critical  to  achieve  Millennium Development  Goals \cite{b2}. Further, the claimed doctor-patient population ratio in India is 1: 1,674 and the numbers are worse in the backward areas\cite{b3} \cite{b4}.
With recent advances in the field of machine learning, technology is being used to solve health care delivery problems. With  this  aim,  we  have  proposed  a model  to  aid  doctors and community health workers with constrained resources and experience  in  decision  making.  An  alarming  low  doctor  to patient  ratio  in  developing  countries  calls  for  an  intervention of  technology  integrated  with  knowledge-based  learning  systems,  especially  in  the  rural  scenario.  This  can  also  ensure that  the  decision  making  is  faster  and  effective  advocating early  diagnosis,  preventive  healthcare,  and  treatment.  A  lot of  attempts  have  been  made  to  use  clinical  decision  support  systems  for  the  prediction  and  classification  of  various diseases  based  on  recorded  patient  data.  Very  less  proven work  has  gone  into  equipping  the  workforce  with  a  tool that  can  predict  the  pathways  of  diagnosis  according  to  the changing symptoms/conditions  of  the  patient.  Therefore,  we have  proposed  a  contextual  care  protocol  which  suggests  the standardised  diagnostic  path  of  treatment  and  medications  to be used by the health workers and doctors with less experience. Presently we have incorporated NICE pathways, NRHM and WHO  guidelines  as  the  base  for  extracting  the  decisions contextually\cite{b5}\cite{b6}. As  a  case  study,  we  have  worked  on  Maternal  Health  and Pregnancy. It  is  reported  that  approximately there  are  210  million  pregnant  women  annually  and  130 million  childbirths  happen.  Out of  this,  around  10  million women  experience  illness. Haemorrhage,  infection,  hypertension,  and  obstructed  labor contribute  to  maternal  illness  and  also  causes  fatality  if  the proper screening is not undertaken. Diseases like malaria, HIV infection,  pneumonia,  tuberculosis,  and  herpes virus  infection can complicate health of the mother \cite{b7}\cite{b8}. The  check-ups  during  pregnancy,  before  birth, is known  as  antenatal  care. Antenatal care is very complex, and thus decision-making tools as an aide come into the picture.

Our goal is to equip Community Health Workers (CHWs) and  the  doctors  alike  in  resource-constrained  settings  with technically driven decision-making tools to improve healthcare service  delivery  at  the  grassroots.  To  achieve  this,  we  have adopted  an  approach  of  weighted  decision  trees. Each of the trees represents one of the possible diseases/care protocol and medication. The weights, with the help of neural networks, are adjusted according to the patient’s current status. However,  if  something  is  erroneous  in  terms  of  questions  or the recommendations, the doctor can give his/her feedback on what  would  have  been  the right  question  or  diagnostic  pathway. Similar feedbacks are clustered and after its investigation, the step is correctly updated.

\section{Related Work}
Machine learning algorithms have been widely used in the medical field to build the disease diagnosis support system. Filippo Amato et al. studied the use of artificial neural networks that can be a powerful tool to help physicians perform diagnosis and other enforcement. He highlights the advantages of ANNs which can process a large amount of data, reduces likelihood of overlooking relevant information and thereby reducing diagnosis time\cite{b9}.\par

Niti Guru et al.  proposed a model based on neural networks for doctors for heart diagnosis. When unknown data is entered, the system generates a list of possible diseases that the patient might suffer from. They also discussed the possibility of using neural networks as an indexing function\cite{b10}. F. S Bakpo et al. presented a framework for diagnosing skin diseases using artificial neural networks\cite{b11}. Shrwan Ram used decision trees and neural networks separately for the classification of Hematology datasets and concluded that Neural networks have better classification performance\cite{b12}.\par

B.N Lakshmi et al. worked out a learning model using C4.5 decision tree algorithm for classification as well as predicting risks induced during pregnancy\cite{b13}. The application proposed by the patent using neural networks identifies important input variables for a medical diagnostic test which was used in training the decision-support systems to guide the development of the tests and helps in accessing the effectiveness of a selected therapeutic protocol\cite{b14}. S.Florence et al. proposed neural network and decision tree algorithm in an integrative manner to predict heart attacks with high amount of accuracy\cite{b15}.\par

Jose´ M. Jerez-Aragone´s proposed a model that combines TDIDT (CIDIM), with a system composed of different neural network topologies to approximate Bayes’ optimal error for the prediction of patient relapse after breast cancer surgery. The CIDIM algorithm selects the most relevant prognostic factors for the accurate prognosis of breast cancer, while the neural networks system takes as input these selected variables in order for it to reach correct classification probability\cite{b16}.\par

L. G. Kabari et al. presented a framework for diagnosing eye diseases using Neural Networks and Decision Trees. They proposed the hybrid model as Neural Networks Decision Trees Eye Disease Diagnosing System (NNDTEDDS) to train younger ophthalmologists. The use of neural networks is in the diagnosis according to the various symptoms, physical eye conditions and that of decision trees is in knowledge extraction from trained neural networks The various rules obtained according to symptoms explains the knowledge acquired in neural networks by learning from previous samples of symptoms and physical eye conditions\cite{b17}.

Neural Network and decision trees are the prominent areas of research that can be found in wide variety of applications in healthcare and diagnosis systems. The above-mentioned works employ these algorithms both individually as well as combined with some specific tasks. The method proposed in this paper is also a combined one, but it is generic and uses a dynamic approach for training and hence it overcomes the shortcomings of existing approaches.
\section{Methodology}
\begin{figure*}[tb]
    \centering
    \includegraphics[scale=0.5]{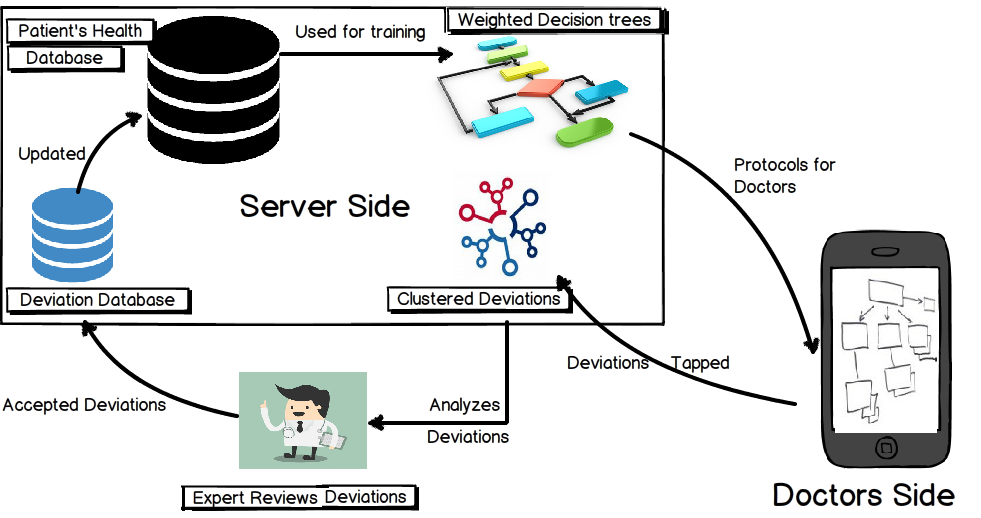}
    \caption{Overview of Model}
    \label{fig:model}
\end{figure*}

Our model, as shown in Fig.\ref{fig:model} has a synthetic dataset which contains a set of diagnosis decision trees and the medical records of the patients under consideration. We use this data to train a neural network to select a diagnosis process. In the case that our model seems to deviate from the optimal decision, the practitioner can report these deviations. These deviations can then be clustered and sent for review to a subject expert and on approval, the deviation will be corrected in our diagnosis tree data. This process continues and optimizes with each update in the data. The following sections describe individual parts of our algorithm.
\subsection{Neural Networks}
We propose to use the medical data to train an Artificial Neural Networks (ANN) classifier. ANN are broadly applied in research because they can model systems in which the relationship between the variables is unknown or very complex. ANN classifier, an example of supervised learning, learns to separate data samples into different classes by finding similar features between the samples of known class. It is a highly non-linear model that has an input layer, hidden layers and an output layer of neurons where each neuron computes a weighted sum of its inputs and interacts with others. The ability to tackle highly complex and non-linear problems give ANN an edge over the other supervised classifiers, especially in healthcare scenarios.
\subsubsection{Input Data}
The purpose of using ANN model is to classify the data to return the list of most probable diseases to let doctors or diagnosticians have a better understanding of patient's current health-related problems to estimate the chances of diseases. The medical record can contain a wide variety of data items. We include:

\begin{itemize}
    \item Meta-data that comprises of case information like visiting date and time, patient information, problems, medications, history etc.
    \item Secondly, we include results from physical examination, laboratory data, symptoms and other information provided by the medical specialist.
\end{itemize}

\subsubsection{ANN model}
The above data, organized and cleaned, are given as the input with each neuron in input layer corresponding to the total number of parameters. We use only one hidden layer here with the number of neurons depending upon input and output layer. The number of hidden neurons should be within the range the neurons in the input layer and output layer for optimal result. For example, if input and output layer has 15 and 25 neurons respectively, we can use around 20 neurons in the hidden layer. Similarly, in the case of 25 neurons in input and only 5 neurons are there in output, the hidden layer can contain around 15 neurons. The ANN model can have one or more hidden layers depending upon the complexity and linearity of data. The output layer consists of one logistic neuron per disease.
\\\\
The dataset is used to train this ANN model which returns the probabilities for new data samples. So that, we can list the most probable diseases for the further treatment process. The accuracy depends on the size of data-set used for training the model. As the model needs to be updated for any medical situation, we train it with a new batch of data samples that are entered every day as data.  We can also adjust the hyperparameters in the model like learning rate and the structure of ANN etc., for better performance and efficiency.
\subsection{Decision Tree and Neural Networks}
\begin{figure}[h!]
    \includegraphics[scale=0.65]{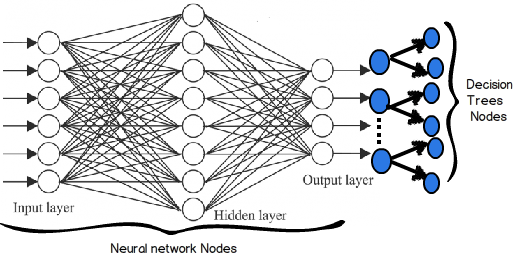}
    \caption{Decision Tree and Neural Network}
    \label{fig:nndt}
\end{figure}

Fig.\ref{fig:nndt} is a representation of weighted decision trees. Using the probabilities given by the ANN described above, we can choose the best decision tree to diagnose a given patient. We choose all trees which clear a certain threshold probability, allowing us to choose and diagnose multiple diseases at the same time.
We then parse the decision trees one node (decision) at a time and recommend tests along the way. This allows us to minimize tests required for a successful diagnosis which would lead to reduced costs for the patient.
\subsection{Decision Trees}
Decision trees are a very widely used format for medical diagnosis and a lot of
human labor has been used to build effective and useful diagnosis trees for
specific diseases. These trees have been tested time and again and have been corrected, optimizing them. Our methodology uses these optimized trees as a base for the care protocol.
\begin{enumerate}
    \item The protocol first involves using the pre-existing information, like age, weight, possible genetic problems, historical problems etc. to first choose the most likely decision tree.
    \item It will start to run through the decision tree at a node by node or question by question basis. However, in order to resolve these nodes, some extra tests may be required.
    \\
    For example, let the first decision tree chosen to be one for diabetes mellitus. The first node of this tree could be asking if the blood sugar level is more than $100ppm$. In order to resolve this, one would need to know the sugar level, knowing which requires a sugar test.
    \item These trees, the order of which is selected by the network will be parsed by recommending tests and will give an optimized diagnosis.
\end{enumerate}
By using pre-built, time-tested and decision trees which are familiar to doctors, we minimize the deployment cost while keeping the doctor with familiar methods. Since these trees would also be quite accurate (as they are human-built), the problem of erroneous trees don't exist.

\subsection{Error Correction and Clustering}
\begin{figure}[h!]
    \includegraphics[scale=0.4]{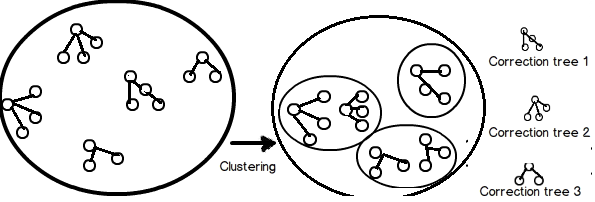}
    \caption{Clustering}
    \label{fig:clus}
\end{figure}

The trees used earlier are time tested and have been optimized manually for a long time. However, it still may be that some corrections and additions may be required for them to work better. Even then, as medical practices change with time, having a set of static decision trees is not desirable. In order to allow our trees to keep up with the changes in the field, we have allowed a doctor side error reporting and correction method. This would work as follows:
\begin{enumerate}
    \item The doctor would see what the current possible diagnoses are, what parameter is being tested right now and what test is being recommended.
    \item The doctor may realize that the test or diagnosis method has become outdated and wants to report an error and give a correction.
    \item The doctor then inputs the correct decision tree in the system, as well as a written description of what went wrong and how is the current tree outdated.
    \item These error corrections will then be received by our system to be reviewed by a human expert/a committee of human experts on whether a given correction is valid or not.
    \item For scalability, we can cluster the error corrections based on the description of the error provided, the corrected decision tree, the doctor's specialization and other parameters to reduce the effort of our experts (Fig. \ref{fig:clus}).
    \item If the corrections are approved, they are added to our repository of decision tree protocols.
\end{enumerate}
This method of reviewed error correction allows our system to keep up with the advances in medicine, allowing us to make the most up-to-date diagnosis methods available, which would lead to better and more accurate diagnosis.\\
Clustering is important as there might be a lot of doctors reposting the same error, and using up our expert's time on repeated error is wasteful. It will allow our experts to focus on errors which are major and are being reported much more often.\\\\

\section*{Conclusion}
Our work describes a realistic and implementable method of contextual care protocol, as this is how doctors diagnose in real life as well. The changing symptoms of a patient can be tracked leading to faster and effective diagnosis. It is also self-correcting as it allows a feedback mechanism with which medical practitioners can suggest changes. It handles the changes in a scalable and verified manner.
This self-correcting nature makes our model better day by day, making it closer to an optimal protocol with time.
\bibliographystyle{IEEEtran}  
\bibliography{conf}  

\begin{thebibliography}{10}
\providecommand{\url}[1]{#1}
\csname url@samestyle\endcsname
\providecommand{\newblock}{\relax}
\providecommand{\bibinfo}[2]{#2}
\providecommand{\BIBentrySTDinterwordspacing}{\spaceskip=0pt\relax}
\providecommand{\BIBentryALTinterwordstretchfactor}{4}
\providecommand{\BIBentryALTinterwordspacing}{\spaceskip=\fontdimen2\font plus
\BIBentryALTinterwordstretchfactor\fontdimen3\font minus
  \fontdimen4\font\relax}
\providecommand{\BIBforeignlanguage}[2]{{%
\expandafter\ifx\csname l@#1\endcsname\relax
\typeout{** WARNING: IEEEtran.bst: No hyphenation pattern has been}%
\typeout{** loaded for the language `#1'. Using the pattern for}%
\typeout{** the default language instead.}%
\else
\language=\csname l@#1\endcsname
\fi
#2}}
\providecommand{\BIBdecl}{\relax}
\BIBdecl

\bibitem{b1}
A.~Mills, R.~Brugha, K.~Hanson, and B.~McPake, ``What can be done about the
  private health sector in low-income countries?'' \emph{Bulletin of the World
  Health Organization}, vol.~80, no.~4, pp. 325--330, 2002.

\bibitem{b2}
G.~Dussault and M.~Franceschini, ``Not enough there, too many here:
  understanding geographical imbalances in the distribution of the health
  workforce,'' \emph{Human Resources for Health}, vol.~4, p.~12, 2006.

\bibitem{b3}
M.~Panda, S.~M. Ali, and S.~Panda, ``Big data in health care: A mobile based
  solution,'' in \emph{International Conference on Big Data Analytics and
  Computational Intelligence (ICBDACI)}.\hskip 1em plus 0.5em minus 0.4em\relax
  IEEE International, March 2017.

\bibitem{b4}
P.~Zurn, M.~R.~D. Poz, B.~Stilwell, and O.~Adams, ``Imbalance in the health
  workforce,'' \emph{Human Resources for Health}, vol.~2, p.~13, 2004.

\bibitem{b5}
\BIBentryALTinterwordspacing
``National institute for health and care excellence (nice).'' [Online].
  Available: \url{https://pathways.nice.org.uk/}
\BIBentrySTDinterwordspacing

\bibitem{b6}
\BIBentryALTinterwordspacing
``National rural health mission - maternal health care.'' [Online]. Available:
  \url{http://nhm.gov.in/nrhmcomponnets/reproductive-child-health/maternalhealth.html}
\BIBentrySTDinterwordspacing

\bibitem{b7}
L.~P. Finnegan, J.~Sheffield, H.~Sanghvi, and M.~Anker, ``Infectious diseases
  and maternal morbidity and mortality,'' 2004.

\bibitem{b8}
B.~N. Lakshmi, T.~S. Indumathi, and N.~Ravi, ``An hybrid approach for
  prediction based health monitoring in pregnant women,'' \emph{Procedia
  Technology}, vol.~24, pp. 1635--1642, 2016.

\bibitem{b9}
F.~Amato, A.~López-Rodríguez, E.~Peña-Méndez, P.~Vaňhara, A.~Hampl, and
  J.~Havel, ``Artificial neural networks in medical diagnosis,'' \emph{J Appl
  Biomed}, vol.~11, pp. 47--58, 12 2013.

\bibitem{b10}
N.~Guru, A.~Dahiya, and N.~Rajpal, ``Decision support system for heart disease
  diagnosis using neural network,'' \emph{Delhi Business Review}, vol.~8,
  no.~1, pp. 99--101, 2007.

\bibitem{b11}
F.~S. Bakpo and L.~G. Kabari, ``Diagnosing skin diseases using an artificial
  neural network, artificial neural networks - methodological advances and
  biomedical applications.''\hskip 1em plus 0.5em minus 0.4em\relax Prof. Kenji
  Suzuki (Ed), InT, 2011, inTech.

\bibitem{b12}
R.~Shrwan, ``Performance evaluation of decision tree and neural networks for
  classification of hematology databases,'' 2015.

\bibitem{b13}
B.~N. Lakshmi, T.~S. Indumathi, and N.~Ravi, ``Prediction based health
  monitoring in pregnant women,'' in \emph{2015 International Conference on
  Applied and Theoretical Computing and Communication Technology (iCATccT)},
  Oct 2015, pp. 594--598.

\bibitem{b14}
\BIBentryALTinterwordspacing
J.~Lapointe and D.~DeSieno, ``Method for selecting medical and biochemical
  diagnostic tests using neural network-related applications,'' 2004. [Online].
  Available: \url{https://www.google.co.in/patents/US6678669}
\BIBentrySTDinterwordspacing

\bibitem{b15}
S.~Florence, N.~G.~B. Amma, G.~Annapoorani, and K.~Malathi, ``Predicting the
  risk of heart attacks using neural network and decision tree,''
  \emph{International Journal of Innovative Research in Computer and
  Communication Engineering}, vol.~2, pp. 7025--7030, 2014.

\bibitem{b16}
J.~M. Jerez-Aragonés, J.~A. Gómez-Ruiz, G.~Ramos-Jiménez, J.~Muñoz-Pérez,
  and E.~Alba-Conejo, ``A combined neural network and decision trees model for
  prognosis of breast cancer relapse,'' \emph{Artificial intelligence in
  medicine}, vol.~27, no.~1, pp. 45--63, 2003.

\bibitem{b17}
L.~G. Kabari and E.~O. Nwachukwu, ``Neural networks and decision trees for eye
  diseases diagnosis,'' \emph{Advances in Expert Systems, Prof. Petrică
  Vizureanu (Ed.)}, 2012, inTech.

\end{thebibliography}

\end{document}